\documentclass[sigconf]{acmart}
\AtBeginDocument{%
  }

\setcopyright{acmlicensed}
\copyrightyear{2018}
\acmYear{2018}
\acmDOI{XXXXXXX.XXXXXXX}
\acmConference[Conference acronym 'XX]{Make sure to enter the correct
  conference title from your rights confirmation email}{June 03--05,
  2018}{Woodstock, NY}
\acmISBN{978-1-4503-XXXX-X/2018/06}

\usepackage{comment}
\usepackage{multirow}

\begin{document}

\setlength{\parindent}{0pt}
\title{Contrastive Representation Learning of Longitudinal Disease Trajectories on Temporal Graphs}

\author{Bastian Pfeifer}
\correspondingauthor
\authornotemark[1]
\email{bastian.pfeifer@medunigraz.at}
\affiliation{%
  \institution{Institute for Medical Informatics, Statistics and Documentation \\ Medical University Graz}
  \city{Graz}
  \country{Austria}
}








\renewcommand{\shortauthors}{Bastian Pfeifer}

\begin{abstract}
  Understanding disease trajectories from longitudinal clinical data remains challenging due to complex temporal dynamics and heterogeneous patient cohorts. Here, we present a contrastive representation learning framework that models multivariate disease trajectories as temporal graphs and learns representations using contrastive graph neural networks. Nodes represent patient observations over time, while edges capture temporal continuity and structural similarity between trajectories. Structure-aware random walks guide contrastive learning to generate embeddings that preserve temporal context and trajectory topology. The resulting representations enable robust clustering of patients with similar disease progression patterns and reveal latent structure in longitudinal data.
\end{abstract}



\received{20 February 2007}
\received[revised]{12 March 2009}
\received[accepted]{5 June 2009}

\maketitle

\section{Introduction}
Longitudinal data arise naturally in numerous application domains, including medicine, biology, epidemiology, and the social sciences, where repeated measurements are collected from the same individuals over time \cite{schussler2019longitudinal, vasaikar2023comprehensive}. A fundamental objective in longitudinal data analysis is to identify groups of subjects with similar temporal behavior, enabling patient stratification, disease subtyping, and a better understanding of underlying progression patterns. 

Traditional longitudinal clustering methods, such as mixed-effects models, latent class models, and functional clustering approaches, have demonstrated success in many applications. However, these methods often rely on restrictive model assumptions and may struggle with  high-dimensional measurements and complex nonlinear dynamics. Moreover, they typically analyze subjects individually and do not explicitly exploit structural relationships between observations across different subjects. Graph neural networks and graph contrastive learning have recently emerged as powerful tools for representation learning on relational data. Nevertheless, their potential for longitudinal data remains relatively unexplored. 

In this work, we propose a novel graph contrastive learning framework for unsupervised longitudinal clustering, called RankWalk. We represent each observation as a graph node and construct a heterogeneous graph by combining temporal edges connecting consecutive observations of the same subject with similarity edges linking subjects exhibiting comparable characteristics at the same measurement time. The resulting node representations are learned using a weighted contrastive objective that emphasizes more informative positive pairs. 

The proposed framework is evaluated on two complementary simulation studies and four real-world longitudinal biomedical datasets. Experimental results demonstrate that the learned representations consistently improve clustering performance compared with existing longitudinal clustering approaches. 

The remainder of this manuscript is organized as follows. Section~\ref{sec:related_work} gives an overview of related work. Section~\ref{sec:proposed_method} introduces the proposed RankWalk framework, including the longitudinal graph construction and graph contrastive learning strategy. Section~\ref{sec:evaluation_strategy} describes the experimental setup and benchmark datasets. Section~\ref{sec:results} presents the simulation studies and real-world benchmark results. Finally, Section~\ref{sec:conclusion} concludes the paper and outlines directions for future research.

\section{Related Work}
\label{sec:related_work}
Longitudinal clustering has been extensively studied using statistical, functional, distance-based, and deep representation learning approaches \cite{lu2025clustering, qiu2025deep}. Functional data analysis methods, particularly those based on functional principal component analysis (FPCA) \cite{wang2016functional,chen2017quantifying}, represent trajectories in a low-dimensional functional basis before clustering in the resulting score space. While effective for smooth trajectories, these methods generally assume continuous underlying functions. Distance-based approaches, such as Dynamic Time Warping (DTW) \cite{giorgino2009computing}, align trajectories by allowing nonlinear temporal distortions and form the basis of clustering algorithms implemented in libraries such as \texttt{tslearn} \cite{tavenard2020tslearn}. Likewise, \texttt{kml3d} \cite{genolini2015kml} extends $k$-means to multivariate longitudinal trajectories using dedicated trajectory distance measures. 

More recently, deep learning methods have been proposed for longitudinal representation learning. These include recurrent LSTM-based architectures and self-supervised approaches such as TS2Vec \cite{yue2022ts2vec}, which learns temporal representations using temporal convolutional networks and hierarchical contrastive learning. VADER \cite{de2019deep,qiu2025deep} combines variational autoencoders with temporal modeling to derive latent representations for clustering. 

Although these methods can capture complex nonlinear temporal patterns, they generally do not explicitly exploit graph-structured relationships between subjects and observations. A graph-based perspective provides a natural framework to integrate temporal dependencies and relationships between observations, enabling the representation of longitudinal data as interconnected trajectories rather than isolated sequences.

Our proposed approach represents longitudinal data as a heterogeneous graph and combines graph neural networks with self-supervised contrastive learning. By jointly modeling temporal dependencies, cross-subject structural similarity, and anchor-guided positive sampling, the proposed framework learns representations specifically tailored to longitudinal clustering without requiring predefined trajectory models, pairwise distance matrices, or labeled data.

\section{Proposed Method}
\label{sec:proposed_method}

\begin{figure*}[h]
  \centering
 \includegraphics[width=\linewidth]{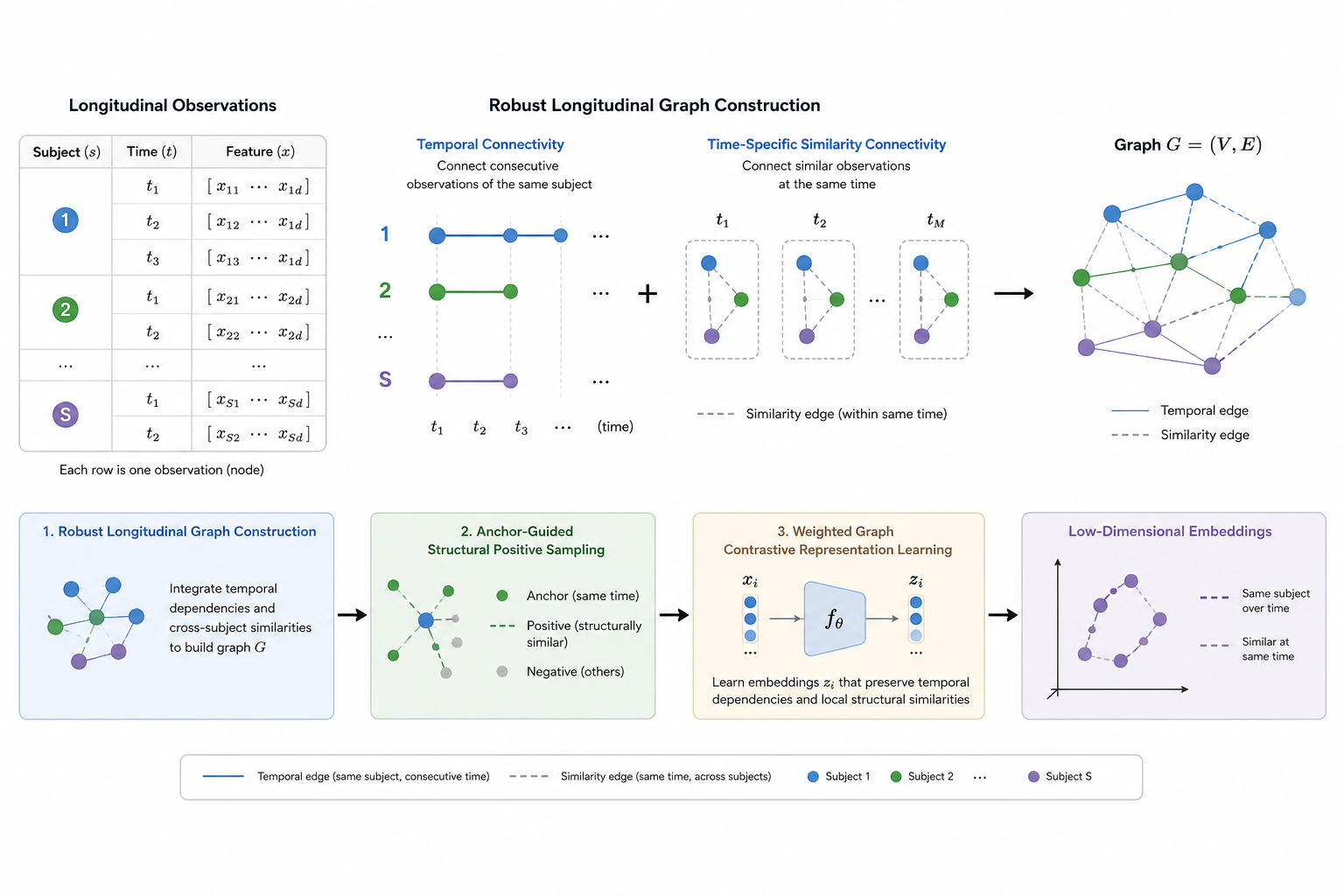}
  \caption{Graphical illustration of the proposed RankWalk method.}
\label{fig:cover_letter}
\end{figure*}

\subsection{Problem Formulation}

Given longitudinal observations collected from multiple subjects over time, we aim to learn low-dimensional representations that preserve temporal evolution patterns and structural relationships among observations.

We define longitudinal data as

\begin{equation}
\mathcal{D}
=
\{(s_i,t_i,\mathbf{x}_i)\}_{i=1}^{N},
\end{equation}

where $N$ denotes the total number of observations. Each observation index $i$ corresponds to one graph node, $s_i$ indicates the associated subject, $t_i$ denotes the acquisition time, and $\mathbf{x}_i\in\mathbb{R}^{d}$ is the observed feature vector.

Each longitudinal observation is represented as a node in a graph

\begin{equation}
G=(V,E),
\end{equation}

where $|V|=N$. The objective is to learn an encoder

\begin{equation}
\mathbf{z}_i=f_{\theta}(\mathbf{x}_i,G),
\end{equation}

which maps each observation into a low-dimensional embedding while preserving
both longitudinal dependencies and local structural similarities.

The proposed framework consists of three main components:
(i) robust longitudinal graph construction,
(ii) anchor-guided structural positive sampling, and
(iii) weighted graph contrastive representation learning.


\subsection{Robust Longitudinal Graph Construction}

Each longitudinal observation corresponds to a graph node. The graph integrates
two complementary types of relationships: temporal dependencies within subjects
and similarity relationships across subjects observed at the same measurement
time.

\subsubsection{Temporal Connectivity}

Temporal edges preserve the evolution trajectory of each subject. For every
subject, consecutive observations are connected:

\begin{equation}
E_{\mathrm{temp}}
=
\{
(v_i,v_j):
s_i=s_j,\;
t_j=\mathrm{next}(t_i)
\}.
\end{equation}

These edges encode longitudinal progression by connecting neighboring time
points within the same individual.

\subsubsection{Time-Specific Similarity Connectivity}

To capture cross-subject similarity, similarity edges are constructed separately
within each measurement time point. Direct distance estimation in
high-dimensional feature spaces may be sensitive to noise; therefore, we employ
a robust rank-aggregation strategy.

Let

\begin{equation}
\mathcal{S}
=
\{S_1,\ldots,S_M\}
\end{equation}

denote randomly sampled feature subspaces. For each subspace $S_m$, the pairwise
distance between two observations acquired at the same time point is computed as

\begin{equation}
d_m(i,j)
=
\|
\mathbf{x}_i[S_m]
-
\mathbf{x}_j[S_m]
\|_2 .
\end{equation}

Instead of averaging raw distances, we aggregate the relative ordering of
distances across subspaces:

\begin{equation}
D_t(i,j)
=
\frac{1}{M}
\sum_{m=1}^{M}
\mathrm{rank}
\left(
d_m(i,j)
\right),
\quad
t_i=t_j=t.
\end{equation}

The rank-based distance reduces sensitivity to individual noisy features and
captures robust neighborhood relationships.

For each node, the $k$ nearest neighbors are selected only among observations
from the same measurement time:

\begin{equation}
E_{\mathrm{sim}}
=
\{
(v_i,v_j):
v_j\in
\mathrm{KNN}_{k}
(D_{t_i}(i,:)),
\;
t_i=t_j
\}.
\end{equation}

The final longitudinal graph is obtained by combining temporal and similarity relations:

\begin{equation}
E
=
E_{\mathrm{temp}}
\cup
E_{\mathrm{sim}} .
\end{equation}

\paragraph{Irregular Measurement Times.}
When observations are acquired at irregular time points, constructing similarity edges only between measurements obtained at identical time instances is no longer feasible. We therefore partition the continuous time axis into a sequence of overlapping sliding windows. 

Let the observation period be divided into windows of equal width with overlap ratio $\rho \in [0,1)$. Each observation is assigned to every window whose temporal interval contains its acquisition time. Within each window, multiple measurements from the same subject are aggregated by averaging the observed feature values, yielding a single representative observation per subject and window. These aggregated observations define window-specific graph nodes, while temporal edges connect consecutive windows belonging to the same subject. Similarity edges are then constructed independently within each window using the same robust subspace rank-aggregation procedure described above. 

The overlapping windows provide smooth transitions between neighboring temporal regions and increase robustness to irregular sampling by allowing observations near window boundaries to contribute to multiple local neighborhoods.


\subsection{Anchor-Guided Structural Sampling}

\subsubsection{Structural Affinity Matrix}

To quantify structural similarity beyond direct graph connectivity, we compute a node affinity matrix using the Jaccard coefficient:

\begin{equation}
J_{ij}
=
\frac{
|\mathcal{N}(i)\cap\mathcal{N}(j)|
}
{
|\mathcal{N}(i)\cup\mathcal{N}(j)|
},
\end{equation}

where $\mathcal{N}(i)$ denotes the neighborhood of node $i$ in the constructed graph.

The resulting matrix

\begin{equation}
J\in\mathbb{R}^{N\times N}
\end{equation}

is used as a structural prior for sampling informative contrastive pairs. It does not modify the graph topology; instead, it guides the exploration process used to identify structurally meaningful positive samples.

\subsubsection{Anchor-Guided Random Walk}

For each anchor node $a$, we perform a stochastic walk

\begin{equation}
v_0=a,v_1,\ldots,v_L .
\end{equation}

At each step, the transition probability from the current node $u$ to a
neighboring node $v$ is defined as

\begin{equation}
P(v|u,a)
=
\frac{
J_{av}+\epsilon
}
{
\sum_{w\in\mathcal{N}(u)}
(J_{aw}+\epsilon)
}.
\end{equation}

Unlike conventional random walks whose transition probabilities depend only on
the current node, the proposed walk explicitly conditions the transition
distribution on the starting anchor node. Therefore, the exploration process is biased toward nodes exhibiting structural similarity to the anchor  \cite{pfeifer2026robust}.

\subsubsection{Rank-Weighted Positive Pair Generation}

During the walk, the first visiting step of each node is recorded  \cite{pfeifer2026robust}:

\begin{equation}
r_a(v)
=
\min\{l:v_l=v\}.
\end{equation}

Nodes discovered earlier are considered more reliable structural matches.
Therefore, the top-$K$ nodes with the smallest visiting ranks are selected as
positive samples:

\begin{equation}
\mathcal{P}_a
=
\mathrm{TopK}_{v}
(-r_a(v)).
\end{equation}

Each anchor-positive pair $(a,j)$ is assigned an importance weight according to
its discovery rank:

\begin{equation}
w_{aj}
=
1-
\frac{
r_a(j)
}
{
\max_{v\in\mathcal{P}_a}r_a(v)+\epsilon
}.
\end{equation}




The collection of weighted positive pairs defines the supervision signal for
contrastive representation learning.


\subsection{Graph Contrastive Representation Learning}

Given the constructed longitudinal graph and sampled positive pairs, we learn
node representations using a graph neural encoder trained with a weighted
contrastive objective.

The initial node representation is obtained through a feature encoder:

\begin{equation}
H^{(0)}
=
\phi(X),
\end{equation}

where $\phi(\cdot)$ denotes a multilayer perceptron.

Graph propagation is performed using graph convolution layers:

\begin{equation}
H^{(l+1)}
=
\mathrm{GCN}
(
H^{(l)},A
).
\end{equation}

A residual connection is applied to preserve the original feature information:

\begin{equation}
H
=
H^{(L)}
+
H^{(0)}.
\end{equation}

The final embedding is produced by a projection head followed by
$\ell_2$ normalization:

\begin{equation}
\mathbf{z}_i
=
\frac{
g(H_i)
}
{
\|g(H_i)\|_2
}.
\end{equation}

\subsubsection{Negative Sampling}

For each positive pair $(i,j)$, a set of negative samples is independently
drawn from the node population:

\begin{equation}
\mathcal{N}_i
=
\{n_1,\ldots,n_M\},
\end{equation}

where each negative node is sampled according to

\begin{equation}
n_m
\sim
\mathrm{Uniform}(V).
\end{equation}

These negative samples represent observations that should not have high
embedding similarity with the anchor node. Therefore, the objective encourages
the model to distinguish structurally meaningful positives from unrelated
observations.

\subsubsection{Weighted InfoNCE Objective}

For a positive pair $(i,j)$, the embedding similarity is defined as

\begin{equation}
s_{ij}
=
\frac{
\mathbf{z}_i^\top\mathbf{z}_j
}
{\tau},
\end{equation}

where $\tau$ is the temperature parameter.

The proposed weighted contrastive objective is

\begin{equation}
\mathcal{L}
=
-
\frac{1}{|\mathcal{P}|}
\sum_{(i,j)\in\mathcal{P}}
w_{ij}
\log
\frac{
\exp(s_{ij})
}
{
\exp(s_{ij})
+
\sum_{n\in\mathcal{N}_i}
\exp(s_{in})
}.
\end{equation}

The weighting mechanism assigns larger contributions to positive pairs that are identified early during anchor-guided structural exploration, reflecting higher confidence in their structural correspondence. The model parameters are optimized by minimizing $\mathcal{L}$ using Adam optimization with gradient clipping.

A graphical illustration of the proposed method is displayed in Figure \ref{fig:cover_letter}.

\section{Evaluation Strategy}
\label{sec:evaluation_strategy}

We evaluated the proposed framework against a range of established methods for longitudinal trajectory representation and clustering using both simulated data and four real-world longitudinal benchmark datasets. The considered approaches included functional principal component analysis (fPCA), dynamic time warping (DTW), k-means for multivariate longitudinal data (kml3d), the self-supervised temporal representation learning approach TS2Vec, and the variational deep embedding approach VaDER. These methods cover classical functional representations, trajectory similarity-based approaches, and deep representation learning techniques.

For all approaches, the resulting trajectory representations or similarity structures were used to derive a subject-level representation, which was then clustered using the same k-means algorithm. This common clustering procedure ensures that performance differences primarily reflect the quality of the learned longitudinal representations rather than differences in the clustering algorithm itself. 

In the simulation studies, the number of clusters for k-means was set to the known number of simulated clusters. The clustering performance was assessed using the adjusted Rand index (ARI), which measures the agreement between the estimated cluster assignments and the known true cluster labels while correcting for chance agreement. Higher ARI values indicate better recovery of the underlying cluster structure. For the real-world benchmark datasets, the resulting clusters were evaluated using survival-based measures, including the concordance index \cite{harrell1982evaluating} and the log-rank test statistic \cite{mantel1966evaluation}.

\subsection{Simulated Longitudinal Data}
To evaluate the performance of the proposed framework under different longitudinal data-generating mechanisms, we conducted two complementary simulation studies. The first simulation represents a classical multivariate longitudinal setting, where clusters are characterized by distinct predefined mean trajectories with correlated outcomes and subject-level temporal variability. 

In contrast, the second simulation considers a more challenging nonlinear dynamic scenario, where cluster structure emerges from latent regime-switching processes, nonlinear temporal dependencies, and transient subject-specific perturbations.

\subsubsection{Simulation 1: Multivariate longitudinal trajectory dynamics}
\label{sec:sim1}

For the first simulation scenario we followed the strategy  described in \cite{zhou2023clustermld}. A multivariate longitudinal model of the form

\begin{equation}
Y_{ij}^{(h)} = f_i^{(h)}(t_{ij}) + r_i(t_{ij}) + \sigma_{ih} + \varepsilon_{ijh},
\end{equation}

\noindent was exploited, where $Y_{ij}^{(h)}$ denotes the $h$-th outcome for subject $i$ at measurement occasion $j$, and $f_i^{(h)}(\cdot)$ represents the $h$-th fixed component or mean trajectory for subject $i$. We specified $H = 5$ outcomes and four underlying clusters. The cluster-specific mean trajectories 
$f^{(h)}(\cdot)$, $h = 1,\ldots,5$, are summarized in Appendix Table~\ref{tab:mean_trajectories} (see Appendix Section \ref{sec:sim1_appendix}) and visualized in Figure \ref{fig:sim1}. The subject-specific random temporal component in equation (12) is modeled by a quadratic random function
\begin{equation}   
r_i(t) = b_{i0} + b_{i1} t + b_{i2} t^2,
\end{equation}

where 

\begin{equation}
(b_{i0},\, b_{i1},\, b_{i2})^\top \sim N\!\left( \mathbf{0},\, \Sigma_b \right).
\end{equation}

\noindent The function \(r_i(t)\) represents a latent subject-level trajectory that is shared across all outcomes \(h = 1,\ldots,5\), thereby inducing correlation among the multivariate longitudinal responses within each subject. In the described setting, the clusters are primarily distinguished by outcomes $Y^{(1)}$ and $Y^{(2)}$, moderately by $Y^{(3)}$ and $Y^{(4)}$, while $Y^{(5)}$ is completely non-informative for cluster separation.



\noindent The random effects $\sigma_{ih}$, 
$h = 1,\ldots,5$, captured correlations among the outcomes and were generated from the multivariate 
normal distribution

\begin{equation}
    \begin{pmatrix}
\sigma_{i1}\\
\sigma_{i2}\\
\sigma_{i3}\\
\sigma_{i4}\\
\sigma_{i5}
\end{pmatrix}
\sim
\text{MVN}\!\left(
0,\;
\Sigma
\right),
\qquad
\Sigma =
\begin{pmatrix}
1 & 0.5 & 0.3 & -0.1 & 0 \\
0.5 & 1 & 0.2 & 0.1 & 0 \\
0.3 & 0.2 & 1 & 0.1 & 0 \\
-0.1 & 0.1 & 0.1 & 1 & 0 \\
0 & 0 & 0 & 0 & 1
\end{pmatrix}
\sigma,
\end{equation}

with $\sigma = \operatorname{diag}(\sigma_{1},\sigma_{2},\sigma_{3},\sigma_{4},\sigma_{5})$. The outcome variable $Y^{(5)}$ served as pure noise and was constructed to be independent of the others.

Measurement errors $\varepsilon_{ijh}$ were sampled from either a normal distribution,
\begin{equation}
\varepsilon_{ijh} \sim \mathcal{N}(0,\eta^{2}).
\end{equation}


The sparse and irregular observation times were generated as follows. For each subject $i$, 
the number of measurements $n_i$ was drawn from a discrete uniform distribution on 
$\{4,\ldots,12\}$. The measurement times $t_{ij}$, $j = 1,\ldots,n_i$, were then obtained by 
sampling $n_i$ random points from the interval $(0,11)$ and taking their order statistics, with the first observation time fixed at $t_{i1} = 0$. In the case of regular observation times, no random sampling was performed; instead, subjects were assigned time-aligned, consecutive measurement times. To ensure adequate temporal spacing, the interval between any two adjacent observation times was constrained to be greater than $0.5$. 
Representative sample trajectories, color-coded by cluster, are shown in Appendix Figure~\ref{fig:sim1} for each outcome (see Appendix Section \ref{sec:sim1_appendix}).

\subsubsection{Simulation 2: Nonlinear regime-switching longitudinal dynamics}
\label{sec:sim2}

The second simulation study was designed to evaluate clustering and representation learning methods under a challenging nonlinear longitudinal setting. In contrast to the first simulation, where cluster differences were induced through predefined mean trajectories, this scenario generated observations from a latent regime-switching process with nonlinear temporal dependencies and transient shocks.

We considered $K=4$ underlying clusters and $H=5$ longitudinal outcomes. For each
subject $i$, observations were generated according to

\begin{equation}
Y_{ij}^{(h)}
=
\mu_{ij}^{(h)}
+
\varepsilon_{ij}^{(h)},
\qquad
h=1,\ldots,H,
\end{equation}

where $\varepsilon_{ij}^{(h)}$ denotes measurement noise,

\begin{equation}
\varepsilon_{ij}^{(h)}
\sim
\mathcal{N}(0,\eta^2),
\end{equation}

with $\eta=0.8$ in the default simulation setting.

The latent mean process was defined as

\begin{equation}
\boldsymbol{\mu}_{ij}
=
\boldsymbol{\alpha}_{k_i,z_{ij}}
+
\boldsymbol{g}
(\mathbf{Y}_{i,j-1})
+
\boldsymbol{\delta}_{ij},
\end{equation}

where $k_i\in\{1,\ldots,K\}$ denotes the cluster membership of subject $i$,
$z_{ij}\in\{1,2,3\}$ denotes the latent dynamic regime at measurement occasion
$j$, $\boldsymbol{\alpha}_{k_i,z_{ij}}$ is a cluster- and regime-specific
effect, $\boldsymbol{g}(\cdot)$ represents a nonlinear autoregressive component,
and $\boldsymbol{\delta}_{ij}$ captures transient shock effects.


\paragraph{Latent regime process.}

The latent state process $\{z_{ij}\}$ was generated according to a
cluster-specific first-order Markov chain:

\begin{equation}
P(z_{ij}=b|z_{i,j-1}=a,k_i=k)
=
\Pi^{(k)}_{ab},
\end{equation}

where $\Pi^{(k)}$ denotes the transition matrix associated with cluster $k$.

The four clusters represented different temporal stability patterns (see Appendix Section \ref{sec:sim2_appendix}).


\paragraph{Cluster- and regime-specific effects.}

The baseline state effects were generated as

\begin{equation}
\boldsymbol{\alpha}_{k,z}
\sim
\mathcal{N}
(
\mathbf{m}_{k,z},
1.2^2 I_H
),
\end{equation}

where the cluster-dependent means were defined as

\begin{equation}
\mathbf{m}_{1,z}=-5,
\qquad
\mathbf{m}_{2,z}=-1,
\qquad
\mathbf{m}_{3,z}=3,
\end{equation}

for all regimes $z$, while the fourth cluster contained regime-dependent
behavior:

\begin{equation}
\mathbf{m}_{4,z}
\in
\{-6,0,6\},
\end{equation}

for regimes $z=1,2,3$, respectively.

This construction induces cluster separation through both average response level
and temporal regime behavior.


\paragraph{Nonlinear temporal dynamics.}

To introduce nonlinear dependencies between consecutive measurements, the
dynamic component was defined recursively as

\begin{equation}
g_h(\mathbf{Y}_{i,j-1})
=
\sin(Y_{i,j-1}^{(h)})
+
0.10
\frac{
(Y_{i,j-1}^{(h)})^2
}
{
1+|Y_{i,j-1}^{(h)}|
}
-
0.10
\tanh
\left(
\sum_{l=1}^{H}
Y_{i,j-1}^{(l)}
\right),
\end{equation}

where $h=1,\ldots,H$.

Therefore, the current observation depends nonlinearly on the complete previous
multivariate state vector, inducing cross-outcome temporal dependencies.


\paragraph{Transient shock effects.}

To simulate abrupt temporal perturbations, each subject was assigned a random
number of shock events:

\begin{equation}
N_i^{\mathrm{shock}}
\sim
\mathrm{Poisson}(0.3).
\end{equation}

Shock times were sampled uniformly from the observation interval:

\begin{equation}
\tau_{ir}
\sim
\mathrm{Unif}(2,T_{\max}).
\end{equation}

For each shock occurring before time $t_{ij}$, the corresponding contribution
was

\begin{equation}
\boldsymbol{\delta}_{ij}
=
\sum_{\tau_{ir}\leq t_{ij}}
0.5
\boldsymbol{\alpha}_{k_i,z_{ij}}
\exp(-(t_{ij}-\tau_{ir})).
\end{equation}

Thus, shocks produced temporary deviations whose influence decayed
exponentially over time.


\paragraph{Irregular observation times.}

To mimic sparse longitudinal sampling, the number of measurements per subject
was generated as

\[
n_i\sim\mathrm{DiscreteUniform}(8,16).
\]

For irregular observation designs, measurement times were sampled from
exponential inter-arrival distributions:

\begin{equation}
t_{ij}
=
\sum_{l=1}^{j}
u_{il},
\qquad
u_{il}\sim\mathrm{Exp}(0.45),
\end{equation}

with observations restricted to the interval $(0,T_{\max})$, where
$T_{\max}=10$. If fewer than five observations were obtained, the subject was assigned an equally spaced observation schedule. For the regular observation scenario, measurement occasions were fixed on an equally spaced grid.

Representative sample mean trajectories, color-coded by cluster, are shown in Appendix Figure~\ref{fig:sim2} for each outcome (see Appendix Section \ref{sec:sim2_appendix}).

\subsection{Real-World Longitudinal Datasets}
To evaluate the proposed framework on real-world longitudinal data, we
consider four benchmark datasets from different biomedical domains:
PBC2, HEART, PAQUID, and AIDS. These datasets represent diverse
longitudinal scenarios, including chronic disease progression, postoperative monitoring, cognitive aging, and infectious disease evolution. They differ in terms of cohort size, number of repeated measurements, observation frequency, and the complexity of the underlying temporal processes. All datasets contain multiple observations per subject collected at irregular time points, resulting in unbalanced longitudinal structures. Details on the datasest can be found in Appendix Section \ref{sec:application_appendix}.

\section{Results and Discussion}
\label{sec:results}

The obtained results from the first simulation demonstrate that the proposed RankWalk approach is highly competitive with established statistical methods for longitudinal clustering (Appendix Figure~\ref{fig:sim1_results}, Appendix Section \ref{sec:results_appendix}). Across the 50 simulation replicates, RankWalk consistently achieves clustering performance comparable to the strongest baseline methods, including fPCA and DTW, while outperforming the deep learning approaches TS2Vec and VaDER. The strong performance of functional principal component analysis (fPCA) is expected, as the simulated trajectories are generated from smooth functional mean curves that closely match the assumptions underlying FPCA. Similarly, DTW benefits from the regular sampling design and the relatively smooth temporal trajectories. In contrast, TS2Vec and VaDER exhibit lower clustering accuracy and substantially higher variability across simulation runs, suggesting that these generic deep representation learning approaches are less well suited to this relatively small stringent functional longitudinal dataset. Overall, these results indicate that the proposed graph-based representation learning framework can achieve performance comparable to specialized longitudinal modeling techniques while providing a flexible representation of temporal and structural relationships among observations.

\begin{figure*}[h]
    \centering
    \includegraphics[width=0.49\linewidth]{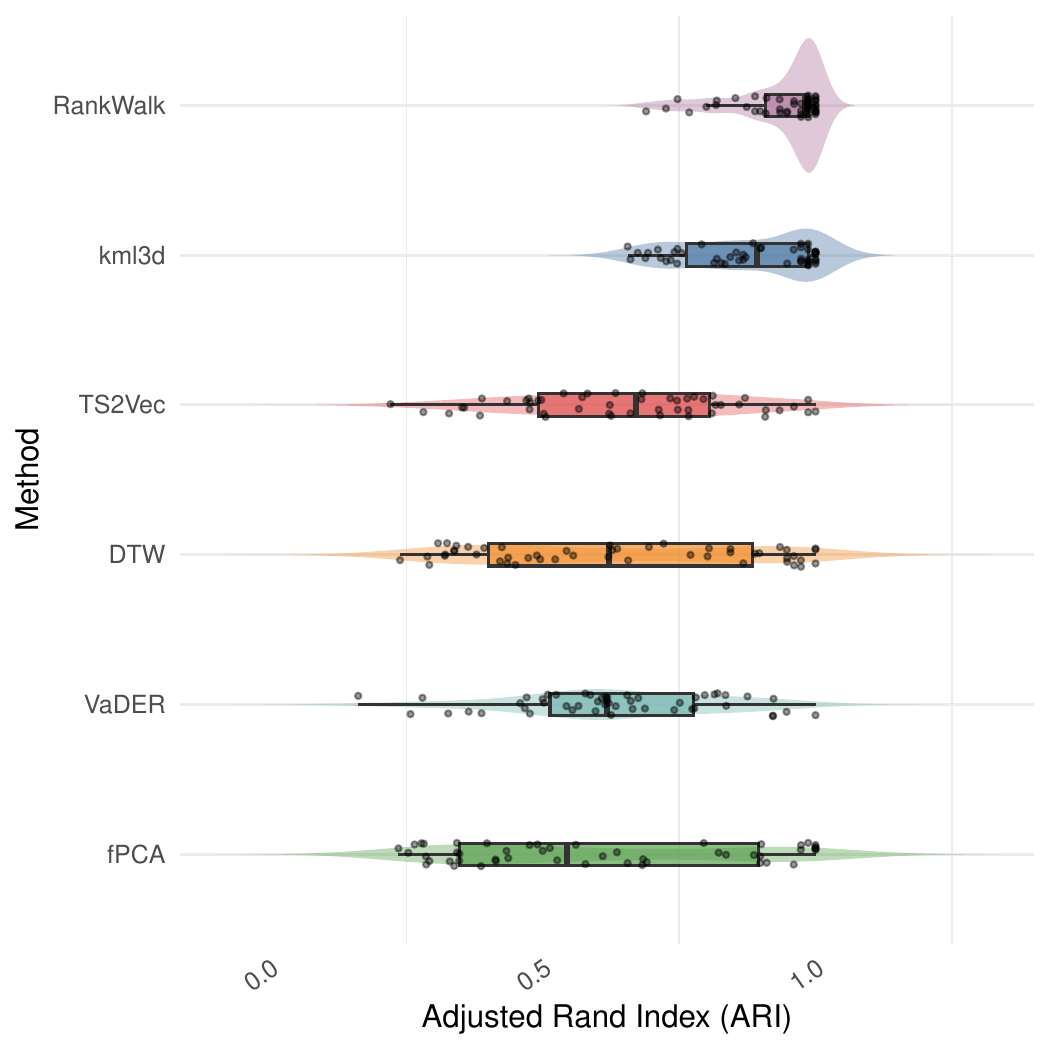}
    \includegraphics[width=0.49\linewidth]{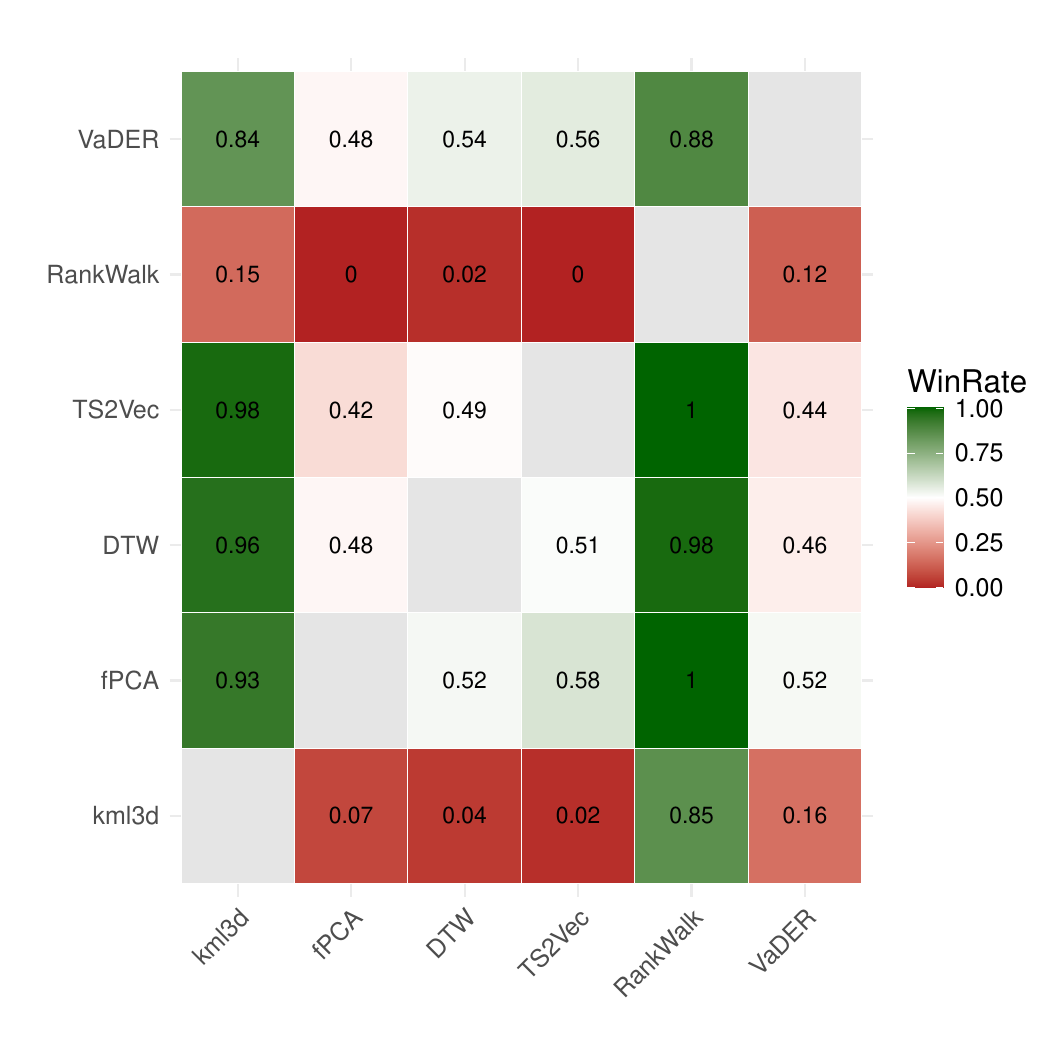}
    \caption{Simulation 1 -- Clustering performance on \textit{noisy} synthetic longitudinal multivariate trajectories (see Section \ref{sec:sim1}). The subject and variable-specific noise levels were set to $\sigma_{i} = 3$, where $i \in \{1,...,5\}$. The global measurement error was set to $\eta=3$. (a) We randomly replaced a single $\sigma_{i}$ with a value out of $[3,20]$ in each iteration. The number of clusters is set to the known ground-truth $K=4$. The results are based on 50 iterations. (b) Pairwise dominance heatmap showing the proportion of simulation replicates in which the method on the $x$-axis achieved a higher ARI than the method on the $y$-axis. Ties are excluded from the calculation, with values close to one indicating consistent superiority across simulation runs.}
    \label{fig:sim1_noise_results}
\end{figure*}

When a randomly selected longitudinal variable is additionally corrupted by noise in each simulation replicate, clear differences in robustness between the competing methods emerge (Figure~\ref{fig:sim1_noise_results}). Since only a subset of the measured outcomes remains informative, methods relying on global trajectory representations become substantially more sensitive to the corrupted variable. This effect is particularly evident for fPCA, whose performance degrades considerably, while DTW also exhibits a noticeable reduction in clustering accuracy. TS2Vec likewise shows limited robustness in this setting. 

In contrast, RankWalk remains remarkably stable and consistently achieves the highest clustering performance across simulation replicates. This robustness stems from the proposed graph construction strategy, where neighborhood relationships are estimated through rank aggregation across multiple randomly sampled feature subspaces. As many subspaces do not contain the corrupted variable, the influence of the noisy measurements is naturally attenuated, resulting in more reliable similarity estimates. Among the competing methods, only kml3d remains consistently competitive, although RankWalk achieves the strongest overall performance. These findings demonstrate that the proposed graph-based representation learning framework is particularly robust when only a subset of longitudinal variables is affected by measurement noise.

\begin{figure*}[h]
    \centering
    \includegraphics[width=0.49\linewidth]{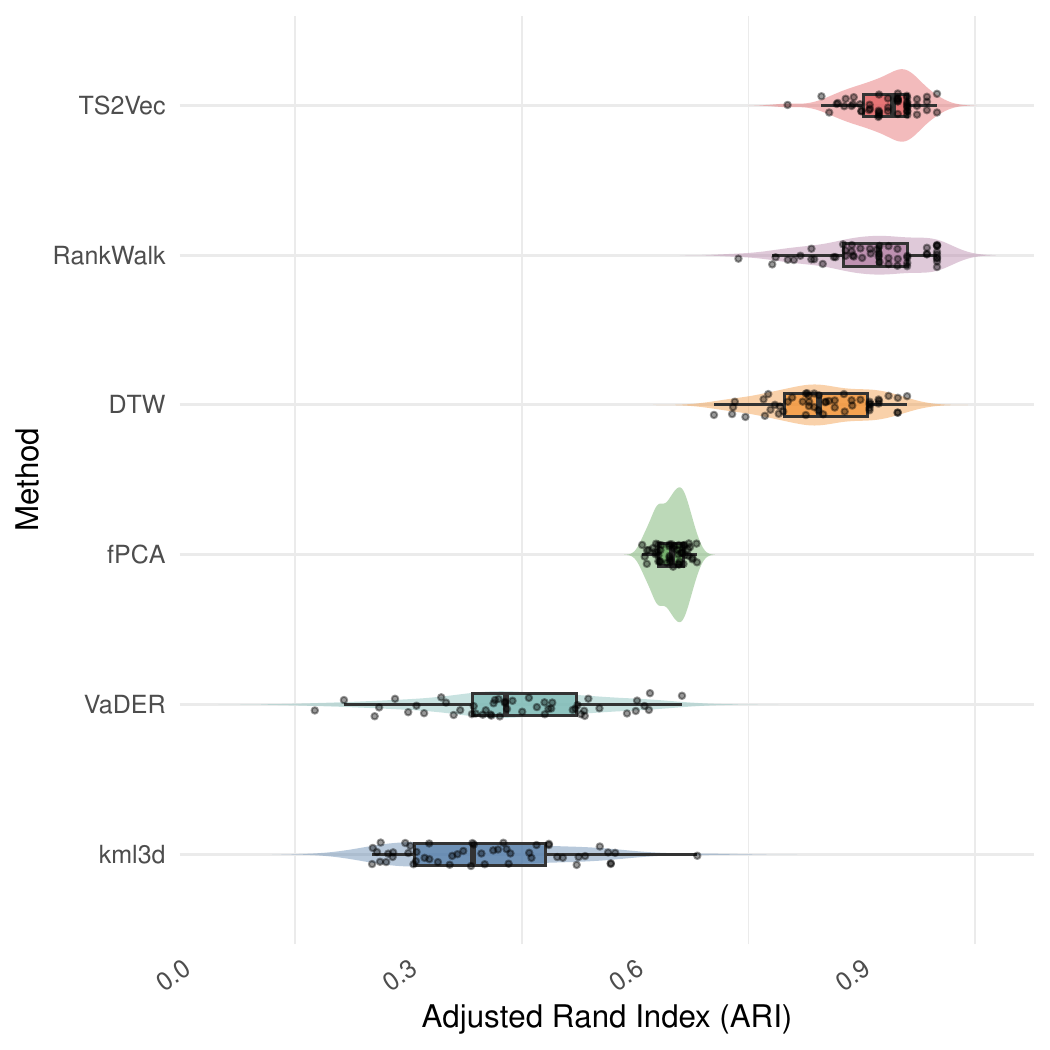}
    \includegraphics[width=0.49\linewidth]{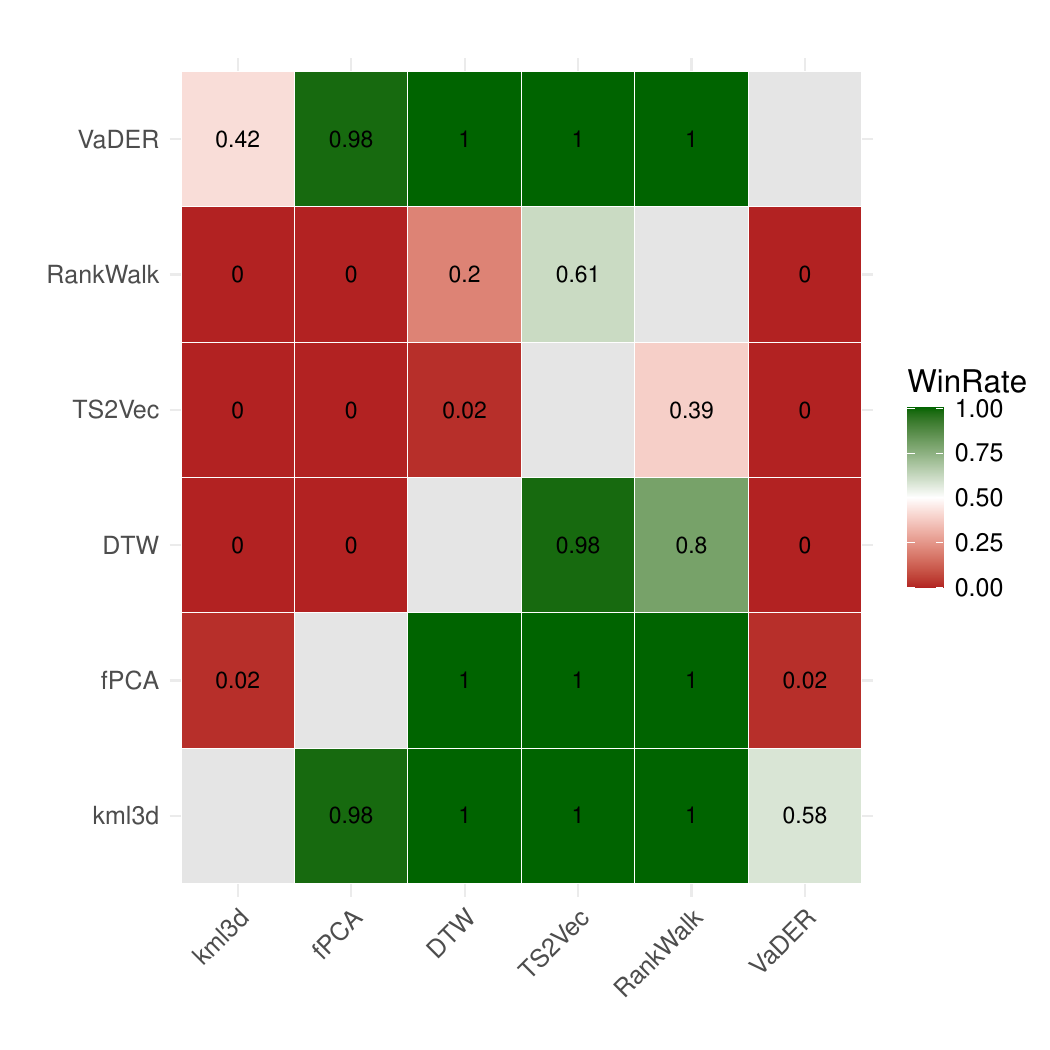}
    \caption{Simulation 2 -- Clustering performance on synthetic longitudinal multivariate trajectories with nonlinear regime-switching dynamics (see Section \ref{sec:sim2}). The number of clusters is set to the known ground-truth $K=4$. The results are based on 50 iterations. (a) Results are shown shown as boxplots. (b) Pairwise dominance heatmap showing the proportion of simulation replicates in which the method on the $x$-axis achieved a higher ARI than the method on the $y$-axis. Ties are excluded from the calculation, with values close to one indicating consistent superiority across simulation runs.}
    \label{fig:sim2_results}
\end{figure*}

The third simulation investigates a more challenging setting characterized by nonlinear temporal dynamics and latent regime transitions. In contrast to the previous experiments, cluster membership is not defined solely by smooth mean trajectories, but by subject-specific dynamic processes generated through cluster-dependent Markov transitions, nonlinear evolution, and transient perturbations. Under this setting, classical trajectory-based approaches show substantially reduced performance (see Figure \ref{fig:sim2_results}). In particular, kml3d performs poorly, indicating that methods relying on predefined longitudinal trajectory structures are not able to adequately capture complex dynamic behavior. Functional PCA achieves moderate performance, but remains limited by its assumption of smooth low-dimensional functional representations.

In contrast, methods designed to capture temporal dependencies, including DTW and TS2Vec, achieve substantially improved clustering accuracy. The proposed RankWalk approach achieves the strongest overall performance and remains highly competitive with TS2Vec, frequently recovering the true cluster structure with near-perfect accuracy. This performance highlights the advantage of integrating temporal connectivity with adaptive similarity relationships in a graph representation. By jointly exploiting within-subject temporal evolution and cross-subject structural similarity, RankWalk can capture complex longitudinal patterns without requiring explicit assumptions about the underlying trajectory shape. These results demonstrate the ability of the proposed framework to model heterogeneous nonlinear longitudinal processes beyond classical functional representations. 

In a subsequent evaluation, we compared the clustering performance of RankWalk on datasets with regular and irregular sampling intervals. As shown in Appendix Figures~\ref{fig:sim1_reg_vs_irreg} and \ref{fig:sim2_reg_vs_irreg}, the proposed sliding-window strategy for transforming irregularly sampled observations into a regular temporal grid performs well, resulting in only a marginal reduction in clustering performance. These findings indicate that the proposed graph construction framework is robust to irregular measurement schedules commonly encountered in real-world longitudinal studies.

In addition, we evaluated the proposed approach on four real-world longitudinal benchmark datasets using survival-based evaluation of the obtained clustering structure. The evaluation was performed using the concordance index (c-index) and the log-rank test statistic. The c-index measures the concordance between the ordering of survival times and the risk ordering induced by cluster assignments, where values closer to one indicate stronger agreement.

To assess the separation between the survival distributions of the identified clusters, we additionally report the $\chi^2$ statistic from the log-rank test. The log-rank test evaluates the null hypothesis that the survival functions are equal across clusters, with larger $\chi^2$ values indicating stronger evidence against this hypothesis. The significance of the test depends on the degrees of freedom, given by the number of clusters minus one. For the four-cluster setting considered in this study, the log-rank statistic
has three degrees of freedom. Accordingly, values exceeding
$\chi^2_{0.95,3}=7.81$ and $\chi^2_{0.99,3}=11.35$ indicate statistically significant separation of survival curves at the 5\% and 1\% significance levels, respectively.

Since longitudinal measurements in real-world cohorts are frequently acquired at irregular time points, we included functional principal component analysis (fPCA) as the main comparator, as it explicitly supports irregularly sampled trajectories. Deep generative approaches such as VaDER were not included in this comparison because they did not show competitive performance in the synthetic experiments and therefore did not provide a reliable baseline for real-world evaluation. Table~\ref{tab:performance} summarizes the results. Across all benchmark datasets, RankWalk achieved higher C-index values and substantially increased log-rank $\chi^2$ statistics compared with fPCA. The largest improvement was observed for the HEART dataset, where RankWalk increased the C-index from $0.57$ to $0.71$ and improved the survival separation statistic from $4.81$ to $52.25$. While the fPCA-based clustering did not reach statistical significance for this cohort, the RankWalk-based clustering resulted in a highly significant separation of survival curves. Similar improvements were observed for PBC2 and AIDS, where both approaches produced significant survival separation, but RankWalk yielded substantially stronger evidence. For PAQUID, RankWalk also improved upon fPCA, although the larger standard deviation of the C-index reflects increased variability across repeated runs.

Overall, these results suggest that representing irregular longitudinal data as graphs, combining within-subject temporal dependencies with cross-subject similarity relationships, enables the identification of trajectory patterns that are more strongly associated with survival outcomes than functional representations based on low-dimensional trajectory bases.

\begin{table}[ht]
\centering
\caption{Comparison of clustering performance across benchmark datasets. Values are reported as mean $\pm$ standard deviation.}
\label{tab:performance}
\begin{tabular}{llcc}
\hline
\textbf{Dataset} & \textbf{Method} & \textbf{c-index} & \textbf{$\chi^2$} \\
\hline
\multirow{2}{*}{PBC2}
  & RankWalk & $0.78 \pm 0.02$ & $246.91 \pm 20.85$ \\
  & fPCA     & $0.76$          & $127.50$ \\
\hline
\multirow{2}{*}{AIDS}
  & RankWalk & $0.69 \pm 0.00$ & $105.86 \pm 6.80$ \\
  & fPCA     & $0.64$ & $53.09$ \\
\hline
\multirow{2}{*}{HEART}
  & RankWalk & $0.71 \pm 0.02$ & $52.25 \pm 7.96$ \\
  & fPCA     & $ 0.57$ & $4.81$ \\
\hline
\multirow{2}{*}{PAQUID}
  & RankWalk & $0.70 \pm 0.02$ & $39.80 \pm 6.64$ \\
  & fPCA     & $0.66$ & $5.06$ \\
\hline
\end{tabular}
\end{table}

\section{Conclusion}
\label{sec:conclusion}

We presented RankWalk, a graph contrastive learning framework for clustering multivariate longitudinal data. By representing longitudinal observations as a heterogeneous graph that jointly captures temporal dependencies within subjects and structural similarities across subjects, the proposed method learns low-dimensional representations tailored to longitudinal clustering. The integration of robust graph construction, anchor-guided structural sampling, and weighted contrastive learning enables RankWalk to identify meaningful trajectory patterns without relying on predefined trajectory models or labeled data. Experimental results on two complementary simulation studies and four real-world biomedical benchmark datasets demonstrated that the proposed framework achieves competitive or superior clustering performance compared with established statistical and deep learning approaches while remaining robust to measurement noise and irregular sampling. Future work will investigate extensions to dynamic graph architectures, missing data mechanisms, and multimodal longitudinal datasets to further improve representation learning in complex biomedical applications.


\section*{Code Availability}
The RankWalk method is available as a Python package on GitHub (\url{https://github.com/pievos101/RankWalk}).

\section*{Acknowledgments}

The authors used OpenAI's ChatGPT (GPT-5.5) to assist with improving the English language and readability of the manuscript. In addition, Figure 1 was created with the assistance of ChatGPT. The authors reviewed and edited all AI-generated content and take full responsibility for the final manuscript.


\section*{Ethics and Privacy Statement}

This paper presents a methodological contribution to longitudinal graph representation learning. No new human subject data were collected, and experiments were conducted using benchmark datasets with appropriate permissions or approvals.

\bibliographystyle{unsrt}
\bibliography{bibliography}


\appendix


\section{Details on Simulated Longitudinal Data}

\subsection{Simulation 1}
\label{sec:sim1_appendix}
See Appendix Table \ref{tab:mean_trajectories}.

\begin{table}[ht]
\centering
\caption{Cluster-specific mean trajectories}
\label{tab:mean_trajectories}
\begin{tabular}{cl}
\hline
\textbf{Outcome $h$} & \textbf{Mean trajectories} \\
\hline

$h=1$ &
\begin{tabular}[t]{l}
cluster 1: $f^{(1)}_{c1}(t) = 8t - 0.6t^{2}$ \\
cluster 2: $f^{(1)}_{c2}(t) = 20 - 6t + 0.3t^{2}$ \\
cluster 3: $f^{(1)}_{c3}(t) = 0$ \\
cluster 4: $f^{(1)}_{c4}(t) = 20$
\end{tabular}
\\[6pt]
\hline
$h=2$ &
\begin{tabular}[t]{l}
cluster 1: $f^{(2)}_{c1}(t) = t$ \\
cluster 2: $f^{(2)}_{c2}(t) = -t$ \\
cluster 3: $f^{(2)}_{c3}(t) = -7t + 0.5t^{2}$ \\
cluster 4: $f^{(2)}_{c4}(t) = -20 + t$
\end{tabular}
\\[6pt]
\hline
$h=3$ &
\begin{tabular}[t]{l}
cluster 1: $f^{(3)}_{c1}(t) = -10 + 6t - 0.4t^{2}$ \\
cluster 2: $f^{(3)}_{c2}(t) = -10 + 6t - 0.4t^{2}$ \\
cluster 3: $f^{(3)}_{c3}(t) = 0.2t$ \\
cluster 4: $f^{(3)}_{c4}(t) = 0.2t$
\end{tabular}
\\[6pt]
\hline
$h=4$ &
\begin{tabular}[t]{l}
cluster 1: $f^{(4)}_{c1}(t) = -1 + t$ \\
cluster 2: $f^{(4)}_{c2}(t) = -1 + t$ \\
cluster 3: $f^{(4)}_{c3}(t) = -1 + t$ \\
cluster 4: $f^{(4)}_{c4}(t) = 10 + 2t - 0.2t^{2}$
\end{tabular}
\\[6pt]
\hline
$h=5$ &
\begin{tabular}[t]{l}
cluster 1: $f^{(5)}_{c1}(t) = -2t + 0.1t^{2}$ \\
cluster 2: $f^{(5)}_{c2}(t) = -2t + 0.1t^{2}$ \\
cluster 3: $f^{(5)}_{c3}(t) = -2t + 0.1t^{2}$ \\
cluster 4: $f^{(5)}_{c4}(t) = -2t + 0.1t^{2}$
\end{tabular}
\\

\hline
\end{tabular}
\end{table}

\begin{figure}[h]
  \centering
 \includegraphics[width=\linewidth]{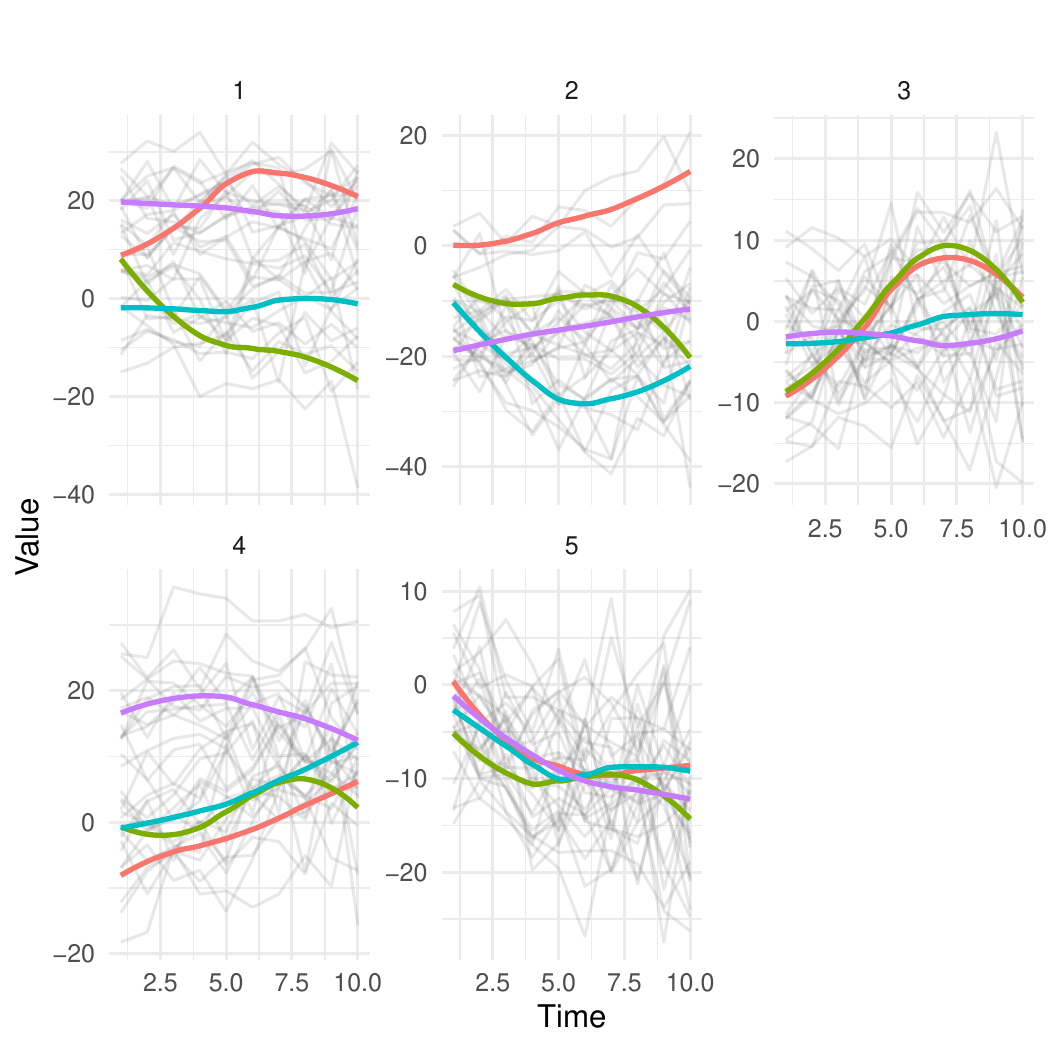}
  \caption{Simulation 1: Longitudinal simulated multivariate data.
   For each longitudinal outcome variable the mean trajectories for each cluster are displayed. See Section \ref{sec:sim1} for details.}
\label{fig:sim1}
\end{figure}

\subsection{Simulation 2}
\label{sec:sim2_appendix}

The four clusters represented different temporal stability patterns:

\begin{equation}
\Pi^{(1)}
=
\begin{pmatrix}
0.93&0.05&0.02\\
0.05&0.90&0.05\\
0.02&0.05&0.93
\end{pmatrix},
\end{equation}

representing a stable regime process,

\begin{equation}
\Pi^{(2)}
=
\begin{pmatrix}
0.82&0.15&0.03\\
0.10&0.80&0.10\\
0.03&0.15&0.82
\end{pmatrix},
\end{equation}

representing a progressive transition pattern,

\begin{equation}
\Pi^{(3)}
=
\begin{pmatrix}
0.75&0.15&0.10\\
0.15&0.70&0.15\\
0.10&0.15&0.75
\end{pmatrix},
\end{equation}

representing an unstable but non-chaotic process, and

\begin{equation}
\Pi^{(4)}
=
\begin{pmatrix}
0.75&0.20&0.05\\
0.20&0.60&0.20\\
0.05&0.20&0.75
\end{pmatrix},
\end{equation}

representing a relapsing-remitting dynamic pattern.

\begin{figure}[h]
  \centering
 \includegraphics[width=\linewidth]{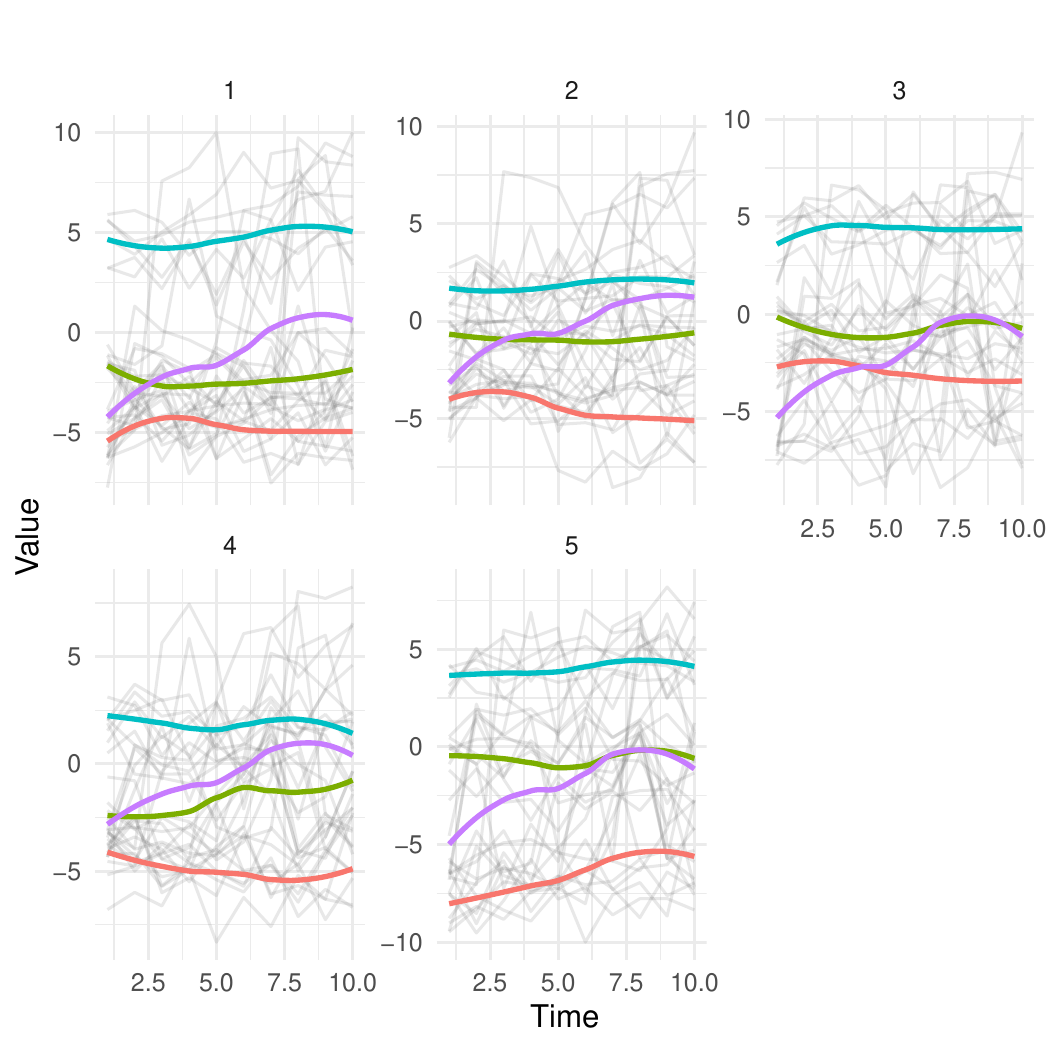}
  \caption{Simulation 2: Nonlinear regime-switching longitudinal dynamics. For each longitudinal outcome variable the mean trajectories for each cluster are displayed. See Section \ref{sec:sim2} for details.}
\label{fig:sim2}
\end{figure}

\section{Details o Real-Worls Data}
\label{sec:application_appendix}

\subsubsection{PBC2}
The PBC2 dataset contains longitudinal observations from patients diagnosed
with primary biliary cirrhosis (PBC), a chronic liver disease characterized by
progressive deterioration of liver function. The data originate from a
clinical study conducted at the Mayo Clinic and contain repeated biochemical
measurements collected during patient follow-up.
The dataset includes 312 patients and a total of 1,945 longitudinal
observations. Each patient contributes a varying number of measurements,
resulting in an unbalanced longitudinal design. The repeated measurements
include several clinical biomarkers related to liver function and disease
severity, such as serum bilirubin, serum cholesterol, albumin, alkaline
phosphatase, SGOT, platelet count, and prothrombin time. In addition to
longitudinal biomarker measurements, the dataset contains survival-related
information, including transplantation and mortality outcomes. The PBC2
dataset therefore represents a setting with heterogeneous disease progression
patterns and irregular follow-up schedules.

\subsubsection{HEART}
The HEART dataset is a longitudinal clinical dataset describing patients
with heart valve disease who were monitored following aortic valve replacement surgery. The dataset was designed for joint modeling studies of longitudinal biomarkers and survival outcomes.
The dataset contains 256 patients with 988 longitudinal observations,
corresponding to approximately 3.86 repeated measurements per patient on
average. Measurements were collected at patient-specific follow-up times,
leading to an unbalanced observation structure. The longitudinal variables
include indicators of cardiac function, such as valve gradient, left
ventricular mass index, and ejection fraction, together with baseline
demographic and clinical characteristics. The dataset captures heterogeneous
postoperative recovery patterns and variability in long-term cardiac
development between patients.

\subsubsection{PAQUID}
The PAQUID dataset originates from a large prospective cohort study investigating cognitive and functional aging in elderly individuals in
France. The study was initiated to characterize age-related changes in
cognitive performance, functional status, and health outcomes over extended
follow-up periods.
The original cohort consists of 3,675 participants followed longitudinally over several years. Participants underwent repeated assessments including cognitive tests, functional measurements, and demographic and health-related evaluations. Due to the long-term population-based design, subjects differ substantially in the number of available follow-up visits and observation intervals. The resulting data structure is highly unbalanced, with slowly evolving individual trajectories reflecting heterogeneous aging processes.

\subsubsection{AIDS}
The AIDS dataset contains longitudinal measurements collected from HIV-infected
individuals enrolled in a clinical study investigating disease progression.
The primary longitudinal marker is the CD4 cell count, which is an important
indicator of immune system status in HIV infection. 
The dataset includes 369 patients and 2,376 longitudinal measurements collected over approximately eight years of follow-up. Each patient contributes between one and twelve repeated measurements, with observation times varying across individuals. In addition to CD4 measurements, the dataset contains baseline covariates and treatment-related information. The AIDS dataset represents a challenging longitudinal setting due to substantial between-patient heterogeneity, irregular measurement schedules, and diverse disease progression trajectories.

\section{Additional Supporting Results}
\label{sec:results_appendix}

\subsection{Results on Simulation 1}
See Appendix Figure \ref{fig:sim1_results}.
\begin{figure*}[h]
    \centering
    \includegraphics[width=0.49\linewidth]{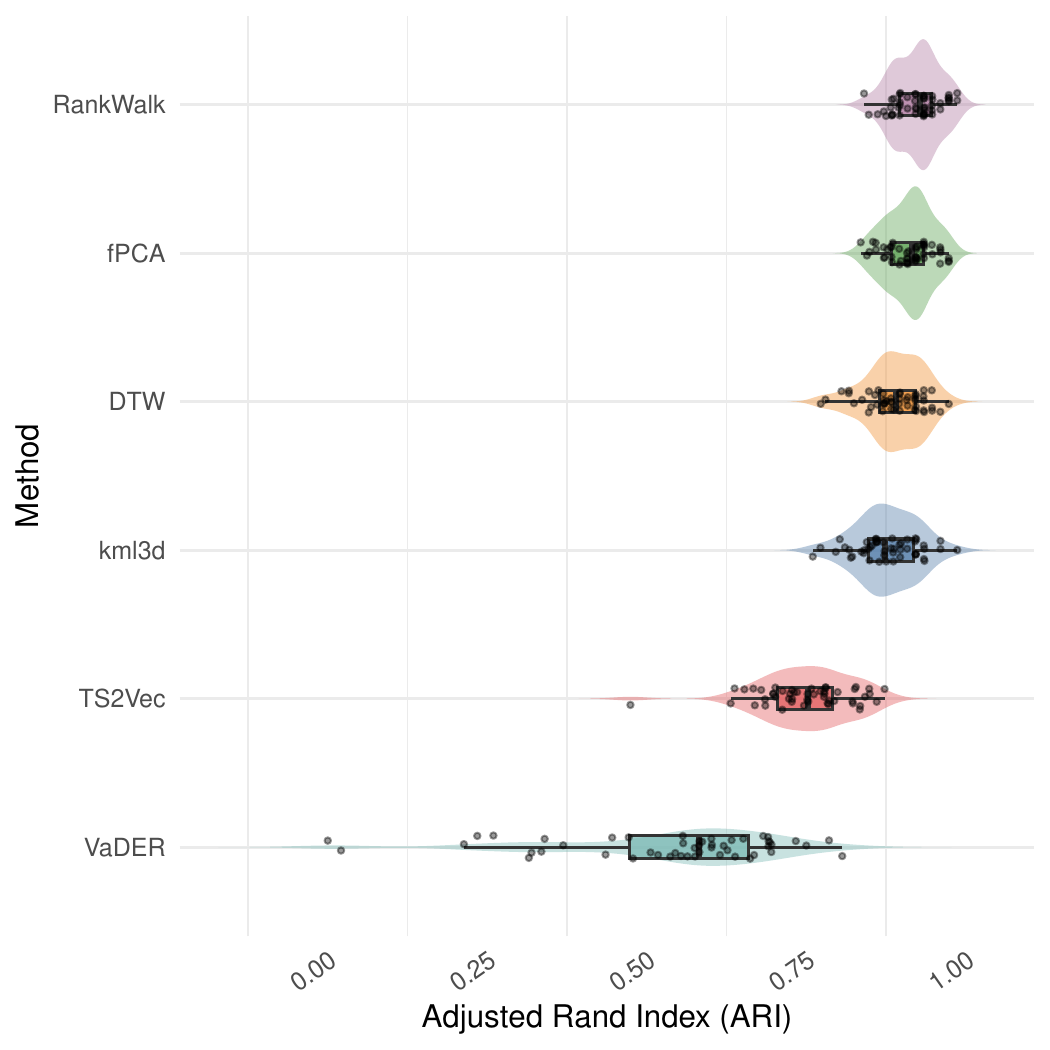}
    \includegraphics[width=0.49\linewidth]{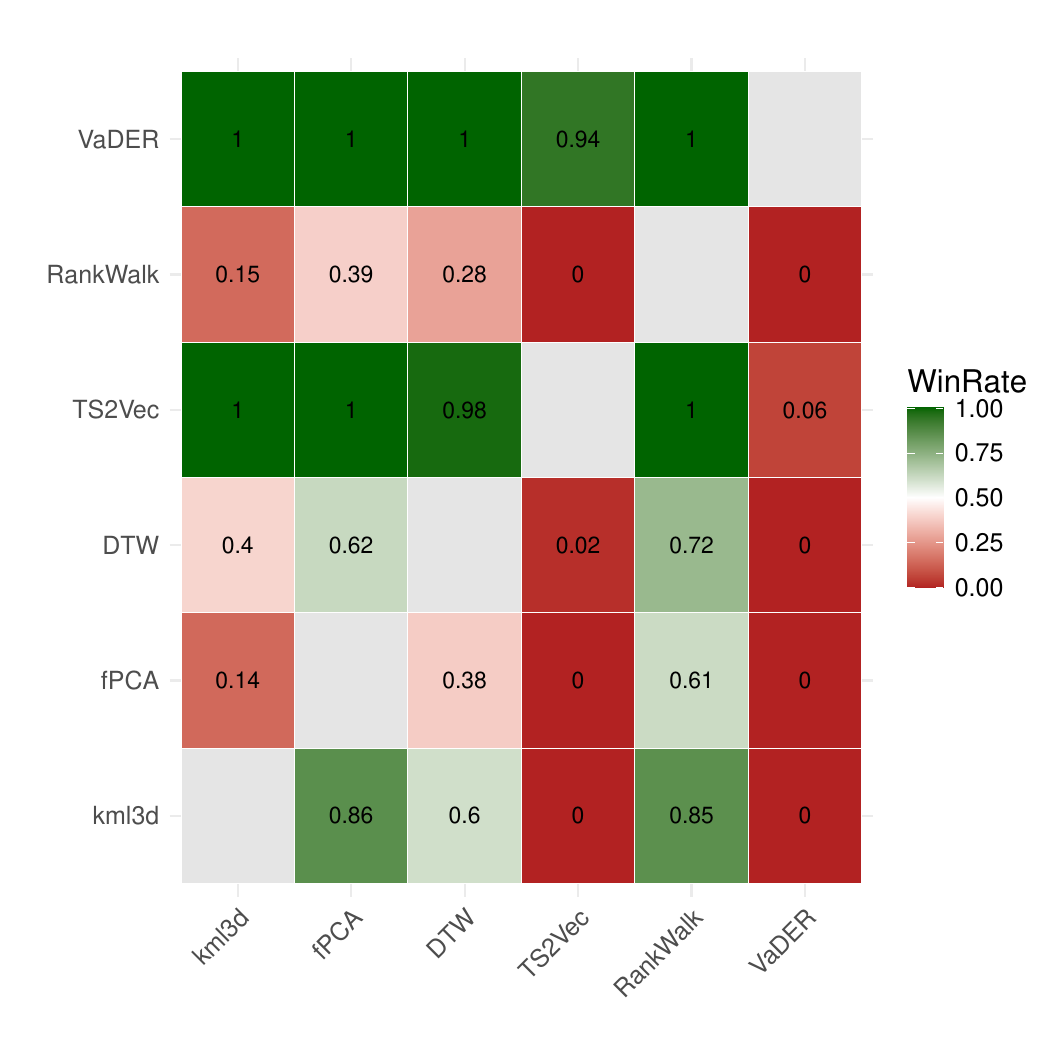}
    \caption{Simulation 1 -- Clustering performance on synthetic longitudinal multivariate trajectories (see Section \ref{sec:sim1}). The subject and variable-specific noise levels were set to $\sigma_{i} = 5$, where $i \in \{1,...,5\}$. The global measurement error was set to $\eta=3$. (a) The number of clusters is set to the known ground-truth $K=4$. The results are based on 50 iterations. (b) Pairwise dominance heatmap showing the proportion of simulation replicates in which the method on the $x$-axis achieved a higher ARI than the method on the $y$-axis. Ties are excluded from the calculation, with values close to one indicating consistent superiority across simulation runs.}
    \label{fig:sim1_results}
\end{figure*}
\clearpage

\subsection{Regular vs Irregular Time Measures}
See Appendix Figure \ref{fig:sim1_reg_vs_irreg} and \ref{fig:sim2_reg_vs_irreg}.

\begin{figure}[h]
    \centering
    \includegraphics[width=0.9\linewidth]{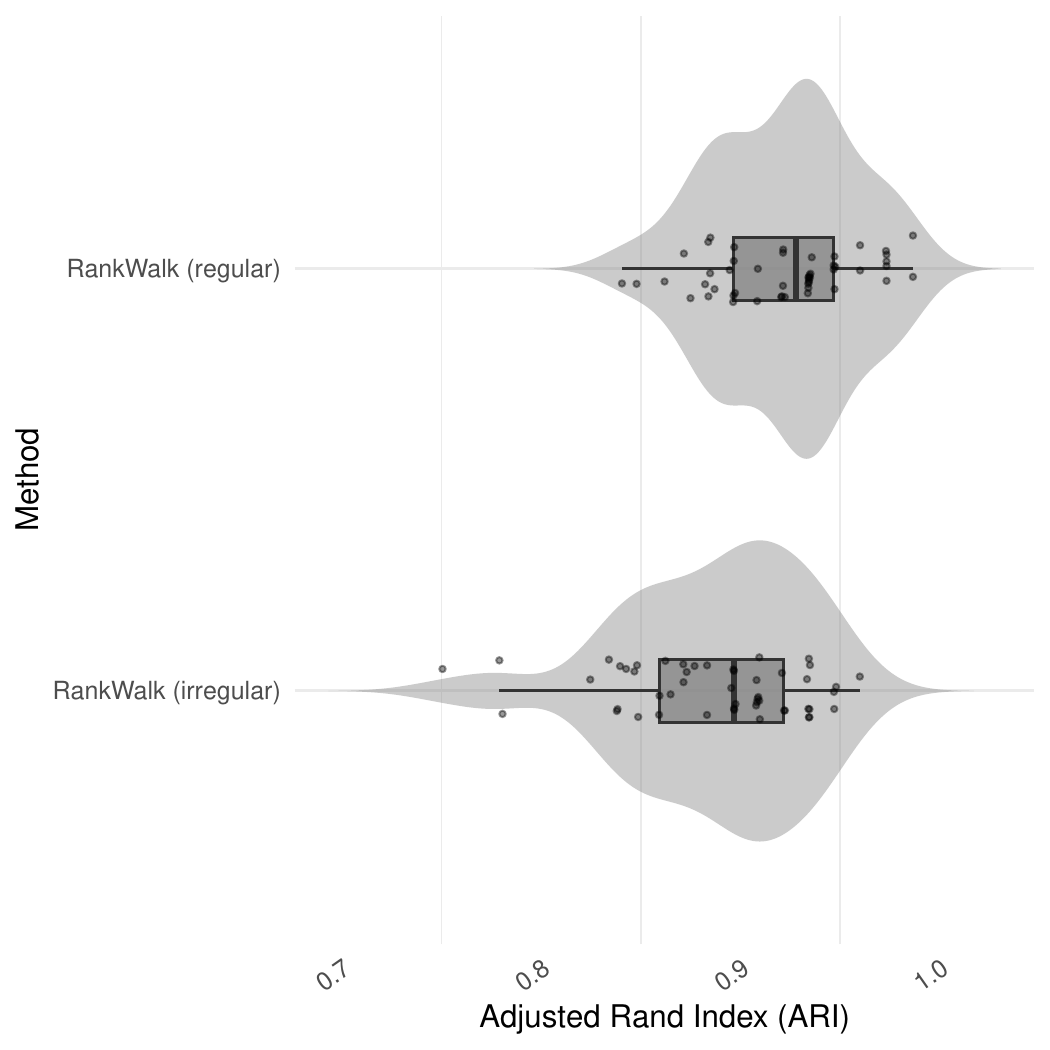}
    \caption{Simulation 1 -- RankWalk on regular versus irregular time measurenments.}
    \label{fig:sim1_reg_vs_irreg}
\end{figure}

\begin{figure}[h]
    \centering
    \includegraphics[width=0.9\linewidth]{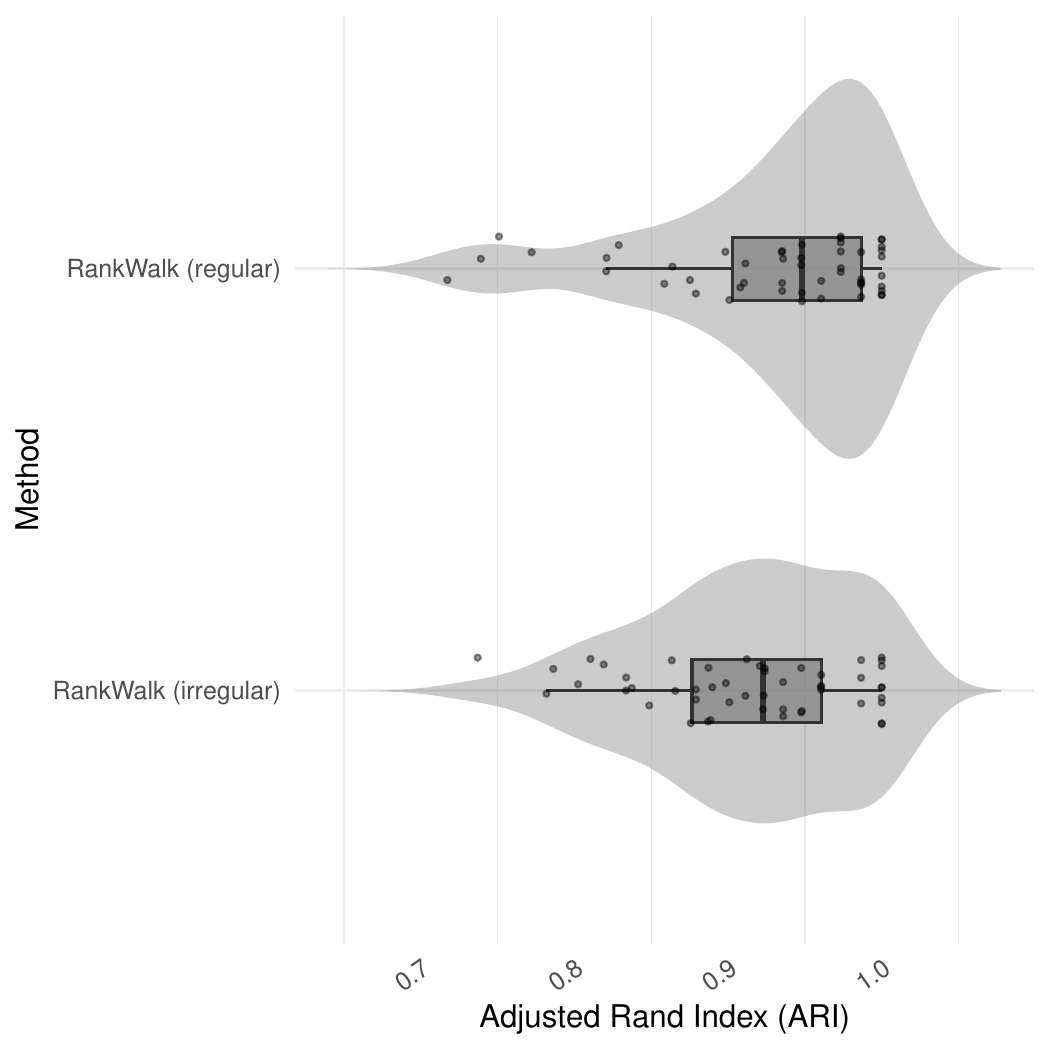}
    \caption{Simulation 2 -- RankWalk on regular versus irregular time measurenments.}
    \label{fig:sim2_reg_vs_irreg}
\end{figure}

\end{document}